\documentclass{article}

\usepackage{arxiv}
\usepackage{graphicx}
\usepackage{amssymb}
\usepackage{amsthm}

\usepackage{amsmath} 
\usepackage{tabularx}
\usepackage{multirow} 
\usepackage{booktabs}

\usepackage[utf8]{inputenc} 
\usepackage[T1]{fontenc}    
\usepackage{hyperref}       
\usepackage{url}            
\usepackage{amsfonts}       
\usepackage{nicefrac}       
\usepackage{microtype}      
\usepackage{lipsum}

\hypersetup{
    colorlinks=true,
    linkcolor=blue,
    filecolor=magenta,      
    urlcolor=cyan,
}
\newcolumntype{Y}{>{\centering\arraybackslash}X}

\title{SILT: Efficient transformer training for inter-lingual inference}

\author{
Javier Huertas-Tato\\
Departamento de Sistemas Informáticos \\
Universidad Politécnica de Madrid \\
\texttt{javier.huertas.tato@upm.es} \\
\And
Alejandro Mart\'in\\
Departamento de Sistemas Informáticos \\
Universidad Politécnica de Madrid \\
\texttt{alejandro.martin@upm.es}\\
\And
David Camacho\\
Departamento de Sistemas Informáticos \\
Universidad Politécnica de Madrid \\
\texttt{david.camacho@upm.es}
}

\begin{document}
\maketitle

\begin{abstract}
The ability of transformers to perform precision tasks such as question answering, Natural Language Inference (NLI) or summarising, have enabled them to be ranked as one of the best paradigm to address Natural Language Processing (NLP) tasks. NLI is one of the best scenarios to test these architectures, due to the knowledge required to understand complex sentences and established relationships between a hypothesis and a premise. Nevertheless, these models suffer from incapacity to generalise to other domains or difficulties to face multilingual and interlingual scenarios. The leading pathway in the literature to address these issues involve designing and training extremely large architectures, which leads to unpredictable behaviours and to establish barriers which impede broad access and fine tuning. In this paper, we propose a new architecture called Siamese Inter-Lingual Transformer (SILT), to efficiently align multilingual embeddings for Natural Language Inference, allowing for unmatched language pairs to be processed. SILT leverages siamese pre-trained multi-lingual transformers with frozen weights where the two input sentences attend each other to later be combined through a matrix alignment method. The experimental results carried out in this paper evidence that SILT allows to reduce drastically the number of trainable parameters while allowing for inter-lingual NLI and achieving state-of-the-art performance on common benchmarks. 

We make our code and dataset available at \hyperlink{https://github.com/jahuerta92/siamese-inter-lingual-transformer}{https://github.com/jahuerta92/siamese-inter-lingual-transformer} .
\end{abstract}




\section{Introduction}
\label{sec:intro}
Natural Language Inference (NLI) has been a recurring issue in Natural Language Processing. In this problem, given two sentences, a premise and a hypothesis, the goal is to determine whether the relation between them can be classified as one of the following categories: entailment, contradiction or neutral. The complexity of this task and its implications in Natural Language Processing (NLP) tasks has resulted in considering NLI as a matter of major importance to build accurate models for general-purpose sentence embeddings generation. 

Inferring the relation between sentences is a challenging task with a wide range of applications. Current approaches, such those based on transformers, deliver decisions based on the semantic content or, in some cases, considering also the context. Models such as BERT or GPT-2 are the greatest exponents of these architectures and have shown excellent skills in varied NLP tasks. However, these models also present limitations, and fail when the task assigned is to consider if there is an entailment relation between a pair of texts. Natural Language Inference has been pointed as a useful instrument to build more accurate language models~\cite{conneau2017supervised}. Through NLI, it is possible to train models and to generate embeddings with state-of-the-art performance but with considerable higher efficiency. Besides, NLI can also be used as an objective during the training phase of language models, helping to generate general-purpose sentences embedding representations~\cite{subramanian2018learning}.

There are plenty of tasks where NLI can serve as extraordinary powerful resource. On the one hand, it is an essential ingredient to better understand natural language. Applications such as question answering~\cite{trivedi2019repurposing}, semantic search~\cite{passaro2020unipi}, automatic summarisation~\cite{pasunuru2018multi} and machine translation~\cite{shimanaka2018ruse} can benefit from NLI, which can also be a powerful method to evaluate models in all these tasks. On the other hand, Natural Language Inference is also an invaluable stand-alone tool with specific objectives, such as paraphrasing detection or facts verification. 

In recent years, there have been notorious advancements in NLI. Architectures such as XLM-R~\cite{conneau2020unsupervised} achieve excellent results in cross-lingual representation learning, improving multilingual BERT by more than 14\% average accuracy on the XNLI dataset. However, the amount of training data and computational resources required to build this model, consisting on 2.5 TB of input data and 500 32GB Nvidia V100 GPUs, represents an extraordinary time-consuming and cost-intensive inefficient process. Progress towards decentralised applications and efficiency requires from usable models that can be deployed and used in real scenarios. Furthermore, despite their outstanding performance on few-shot and zero-shot tasks, models such as XLM-R never report on their inter-lingual capabilities. Given this, and to the best of the authors knowledge, inter-lingual classification has never been attempted in the context of NLI.  

In this paper, we present SILT, a Siamese Inter-Lingual Transformer architecture for Natural Language Inference. It involves a total change towards efficiency, requiring just one GPU to provide state-of-the-art results in two datasets of 3.4 MB and 209.9 MB respectively. This entails a reduction of reduction of 92\% of trainable parameters, from 280M required by XLM-R, to 20M required by SLM. More specifically, this research includes three main contributions:

\begin{enumerate}
    
    \item The definition of SILT, a novel Siamese Inter-Lingual architecture for embedding alignment that crosses mixed language sentence pairs to determine if there is entailment.
    
    \item A Spanish version of the SICK dataset called SICK-es, that can help to train and evaluate multilingual language models.
    
    \item A thorough evaluation of the model in two state-of-the-art datasets, SICK and MNILI + XNLI, and a comparison against state-of-the-art approaches, including our inter-lingual results.

\end{enumerate}

Finally, the remaining of this article is organised as follows: Section~\ref{sec:sota} provides an insight of the state-of-the-art literature related to this research. Section~\ref{sec:align} presents the SILT architecture. Section~\ref{sec:setup} includes information about the experimental setup and Section~\ref{sec:results} presents a series of conclusions an future work lines.

\section{Related work}
\label{sec:sota}

Natural Language Processing, despite being a research field largely studied, is increasingly gaining importance due to new techniques based on Deep Learning and Transformers. Models such as BERT~\cite{kentonbert}, GPT~\cite{radford2018improving} or XLM~\cite{radford2018improving} have proved to be able to answer questions~\cite{qu2019bert}, summarise texts~\cite{liu2019text} or perform natural language inference~\cite{yang2019enhancing} with impressive results. Nevertheless, there is still a long way to go. The sheer size of some architectures or the poor performance when applied to different domains are some issues that evidence the need for further research. In the next subsections, we describe the state-of-the-art literature on Natural Language Inference, the task we face on this research and how it is being addressed with the most recent NLP models.

\subsection{Natural Language Inference}
\label{ssec:nli}

Natural Language Inference remains as a complex and interesting task with a lot of potential in NLP. It can be viewed as an essential step towards human language understanding. Considering if a pair of sentences have an entailment relation, or not, involves a great deal of knowledge, and this is what makes NLI so interesting. 

A recent research highlights the great significance of Natural Language Inference in NLP tasks~\cite{conneau2017supervised}. The authors, in fact, compare this relation with the ImageNet dataset and its widely used to pre-train models for image classification tasks. The researchers demonstrate how this task can be exploited to build generic sentence encoders, changing the classical unsupervised approach and adopting a supervised one. In NLI, the baseline dataset is the Stanford Natural Language Inference corpus (SNLI)~\cite{bowman2015large}, consisting of 570 thousand pairs of sentences labeled for entailment, contradiction or neutral. 

Recent literature devoted to tackle the NLI problem takes the SNLI corpus as basis and address specific issues surrounding this issue, such as risk of bias, lack of explainability or high dependence on the language. Most approaches leverage deep architectures, founding recurrent neural networks as the most profitable architecture. Chen et al.~\cite{chen2017enhanced} follow this path and employ LSTMs. These authors emphasise the importance of exploring the syntax and argue that recurrent networks achieve high accuracy when syntactic parsing information is included. The networks proposed consists of three components: a first input encoding step that includes a bidirectional LSTM and a Tree-LSTM, which allows to collect both context and linguistic information, a modelling step based on local inference and a soft alignment process and finally a inference composition step using again bidirectional LSTM. This complex architecture achieves 88\% accuracy on SNLI. In this work, we introduce two main improvements. First, we replace LSTMs with Multi-head self-attention and pre-trained transformers to avoid RNN shortcomings. Secondly, the architecture allows sequences of different length to be compared without padding.

The increasing complexity of the architectures used to tackle NLI tasks has also given rise to research focusing on lighter options. As an example, Parikh et al.~\cite{parikh2016decomposable} focus on the simplicity of the architecture to achieve state-of-the-art results in the same dataset with an order of magnitude lower parameters. The authors expose that extracting and aligning small local structures can help to perform NLI tasks. This model, however, does not allow a multilingual approach. The authors employ soft alignment to create a series of small inference problems that are later combined. The alignment problem has also been largely studied in the state-of-the-art literature. Other researchers propose a new model for NLI which includes alignment factorization layers~\cite{tay2018compare}. This allows to reduce the alignment vectors and to build a compact architecture with higher interpretability.

It is also worth mentioning that researchers have found that NLI corpus such as SNLI include hidden information that can reveal the label in 67\% of the hypothesis by just looking at it. This is due to the approach used to generate these corpus, where workers, once given a sentence, are asked to generate three sentences with entailment, neutral and contradiction relation. The authors evaluate current NLI models and reveal that the accuracy decreases when the evaluation is performed on the remaining clean hypothesis. This risk of bias has also been pointed by other researchers. In 2019, a research focused on this problem and calculate the probability of a premise according to a hypothesis~\cite{belinkov2019don}. The results demonstrate that it is possible to reduce the bias and build more robust models. In a similar research~\cite{he2019unlearn}, the authors attack the bias that exists on NLI datasets, arguing that a negation is usually associated with a contradiction label. The approach is based on training a biased model to later train a non-biased model using examples that cannot be classified correctly according to the bias. 

There is also a problem that has raised in relation with NLI methods. Although there are recent approaches that obtain high accuracy rates in data corpus such as SNLI, they lack on explainability capacities. In certain cases, it is required not only to make a decision an establish whether there is an entailment relation (or not) between the hypothesis and the premise, but to justify that decision. Camburu et al.~\cite{camburu2018snli} extend the SNLI dataset by including human explanation and show that it can be used to train new models able to provide justifications. Other authors~\cite{kumar2020nile}, also pursue to increase the explainability of the models, showing that it is possible to generate faithful explainability of every decision.

Although Natural Language Inference is repeatedly pointed as an important task in a variety of NLP tasks, its application to specific domain requires adaptation. For instance, clinical domains involve high complexity, and general-purpose datasets such as SNLI are not adequate. Romanov and Shivade~\cite{romanov2018lessons} release the MedNLI dataset, starting from the SNLI dataset and including annotations by doctors that involve domain knowledge. The use external knowledge has also been studied by Chen et al.~\cite{chen2018neural}, demonstrating that neural language models benefit from this information and the training data may not be sufficient to accomplish complex tasks. This is also the case of dialogue agents~\cite{welleck2019dialogue}.

Other possible improvement point lies in the sentence representation scheme, which infer different characteristics and types of reasoning from the text. A research collects different NLI datasets in order to provide the scientific community with a valuable tool to evaluate natural language understanding models and their capacity to capture types of reasoning~\cite{poliak2018collecting}.

Finally, it also important to mention that every language involves essential information to build models with high language understanding capacity. Due to this, different researchers have worked on building corpus on different languages, such as Persian~\cite{amirkhani2020farstail} or Chinese~\cite{hu2020ocnli}. These two example demonstrate the need for multilingual approaches, able to deal with Natural Language Inference tasks no matter the language is. Reaching an inter-lingual solution is a mandatory point in this work.

\subsection{Transformers}
\label{ssec:trans}

BERT~\cite{kentonbert} has become one of the greatest exponents of transformers. By pre-training these bidirectional architectures on unlabeled data and taking advantage of the near context, it is possible to build base models that only require one more specific output layer to perform a specific task. All the previous layers remain intact, which helps to follow use these architectures in a plethora of domains. The authors focused on two tasks: next sentence prediction and mask terms prediction, but they also show the performance in other scenarios, including the SNLI dataset. The Transformers library~\cite{Wolf_Debut_Sanh_Chaumond_Delangue_Moi_Cistac_Rault_Louf_Funtowicz_2019} consists of different pre-trained models facilitating their use and adaptation and has become the principal source of this type of models.

Different researches have investigated the limitations of BERT. In NLI tasks, this Bidirectional Encode can find difficulties due to the need of specific knowledge of the domain~\cite{yang2019enhancing}. As shown in previous section with general architectures for NLI, researchers have also evaluated the use of external knowledge in combination with BERT. According to this approach improves the performance of the model when the knowledge is used to adjust the global inference layer. Other researchers explore the knowledge stored in the Intermediate layers of BERT models, in contrast to typical approaches where only the last layer is used~\cite{song2020utilizing}. Through a fine tuning process with specific pooling strategies and an additional pooling module, it is demonstrated that improving internal layers leads to better models with higher generalisation capacity. The approach is tested in two different domains, natural language inference but also aspect based sentiment analysis. 

The great capacity of Transformers to deal with NLP tasks has led to position these models as the main reference in the state-of-the-art literature. However, in contrast to other domains, evaluating these models involves certain difficulties. This problem has been studied by Desai and Durret~\cite{desai2020calibration}, showing that models such as BERT or RoBERTa increase error rates when using in different domains. In general, Transformers are heavily dependent on the domain, and require specific fine-tuning to achieve good results in a particular domain. In addition to fine-tuning, and like traditional NLP approaches, Transformers also benefit from external human knowledge to improve the results~\cite{yang2019enhancing}. The authors test this approach in two NLI datasets, MultiNLI and Glockner, showing that integrating external knowledge allows to improve the performance in both corpus. Transformers have also been evaluated to generate sentence embeddings, while a stochastic answer network architecture is used to lead the inference process~\cite{liu2018stochastic}. The main novelty of this work lies in the multi-step procedure employed which, according to the authors, allows to extract more complex inferences. 

In general, research has largely evaluated the performance of these architectures. Despite the performance of these models in a large variety of tasks, there are, however, still behaviours and low results~\cite{aspillaga2020stress} in specific tasks that encourage researchers to further study on them. On top of these flaws, inter-lingual capabilities of models are not reported, always resorting to language-matched textual pairs, instead of mixed-language pairs.

        \begin{figure*}  
    		\centering
    		\includegraphics[width=0.9\textwidth]{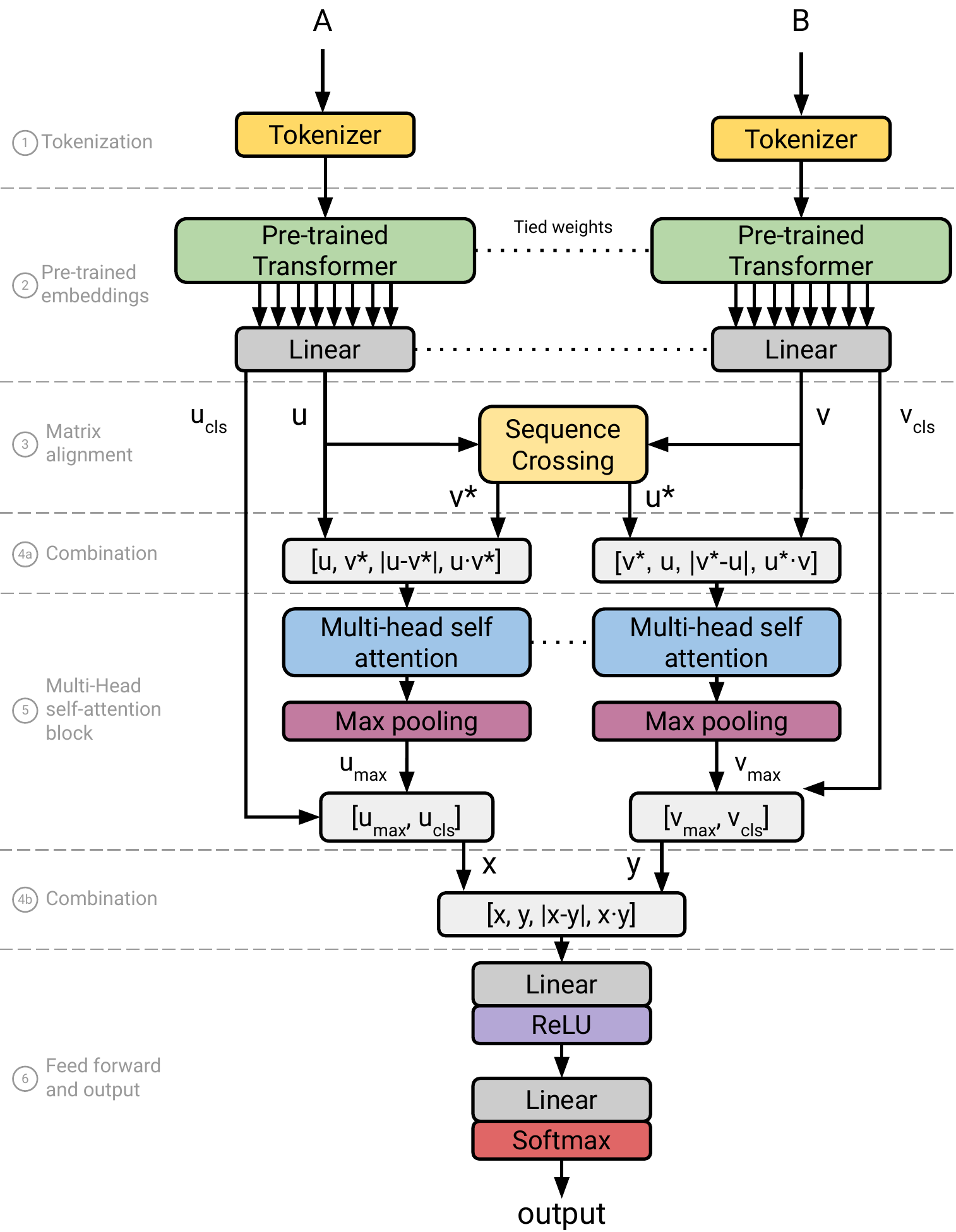}
    		\caption{Topology of the embedding alignment approach.}
    		\label{fig:1_topology}
    	\end{figure*}

\section{The SILT architecture}
\label{sec:align}    
    The Siamese Inter-Lingual Transformer (SILT) involves two siamese multi-lingual transformers, a matrix alignment method and a multi-head self-attention block. This architecture allows to efficiently achieve inter-language inference, classifying pairs of matched and unmatched language textual pairs. While typical NLI approaches train the whole architecture, in this research we demonstrate that pre-trained transformers entail sufficient initial knowledge in order to build efficient architectures for natural language inference tasks. The SILT architecture eliminates the need for fine-tuning the pre-trained transformers by adding a matrix alignment process followed by a multi-head self-attention block that is applied to each input sequence separately. 
    
    Fig.~\ref{fig:0_matrix_alignment} shows a schematic representation of the SILT architecture. As can be seen, the two input sentences pass through the two pre-trained siamese transformers to obtain the embeddings at every layer level, which are then linearly combined into a lower dimension. Then, the projected embeddings are crossed through a matrix alignment operation, a modified alignment from ESIM~\cite{chen2017enhanced}, where the matrix product is attended by each projected embedding. The aligned matrices are interpreted via siamese multi-head self-attention blocks. A max pooling operation turns each sequence to a vector. Each output vector of the self-attention block is combined with the corresponding projected $cls$ token of each sentence, in order to keep information of the original embedding at this step. Finally, both branches of the model are combined via a fully connected feed-forward layer with ReLU activation.
    
    
    The SILT architecture follows a multi-step process:
    
\begin{itemize}
     
    \item     \textbf{Step 1: Tokenization.} It is performed over the non-preprocessed sentence. Most tokenization methods for multi-lingual sentence embeddings work at a character level, and often supports casing, which is our case. An attention mask is provided for the transformer, where $1$ is given for any character and $0$ is used for padding. SILT uses the tokenizer of the pre-trained transformer.
        
    \item    \textbf{Step 2: Pre-trained embeddings.} Transformer models take a tokenized sentence and return a group of word embeddings, along the $cls$ token which is often used to represent the whole sentence. We use the transformers as feature extractors, having their weights frozen and shared. All hidden states of the transformer are extracted, each one producing an embedding with dimension $D$. 
        
    The combined length of all hidden states of the transformer would be too computationally expensive for most operations presented here. Thus, in order to reduce dimensionality, all hidden states of the transformer are combined with the same linear projection to produce the combined embeddings $u$ and $v$ for sentences $A$ and $B$ respectively. Furthermore, this avoids to discard redundant features extracted from the tokenized sentence.
        
    \item     \textbf{Step 3: Matrix alignment.} Both word embeddings $u$ and $v$ are aligned via the method displayed in Fig. \ref{fig:0_matrix_alignment}. Each sentence is represented via a sequence of word embeddings of dim $D$, and a sequence length of $L_u$ and $L_v$ for $u$ and $v$ respectively. From both word sequences we get a similarity matrix $S$. Applying a softmax activation to $S$ and multiplying the result by $v$ we obtain $u^*$, which means u informed by v. To get both $u^*$ and $v^*$, equations~\ref{eq:u_star} and~\ref{eq:v_star} are used respectively.
    
        \label{ssec:matrixalign}
         \begin{figure*}  
    		\centering
    		\includegraphics[width=0.9\textwidth]{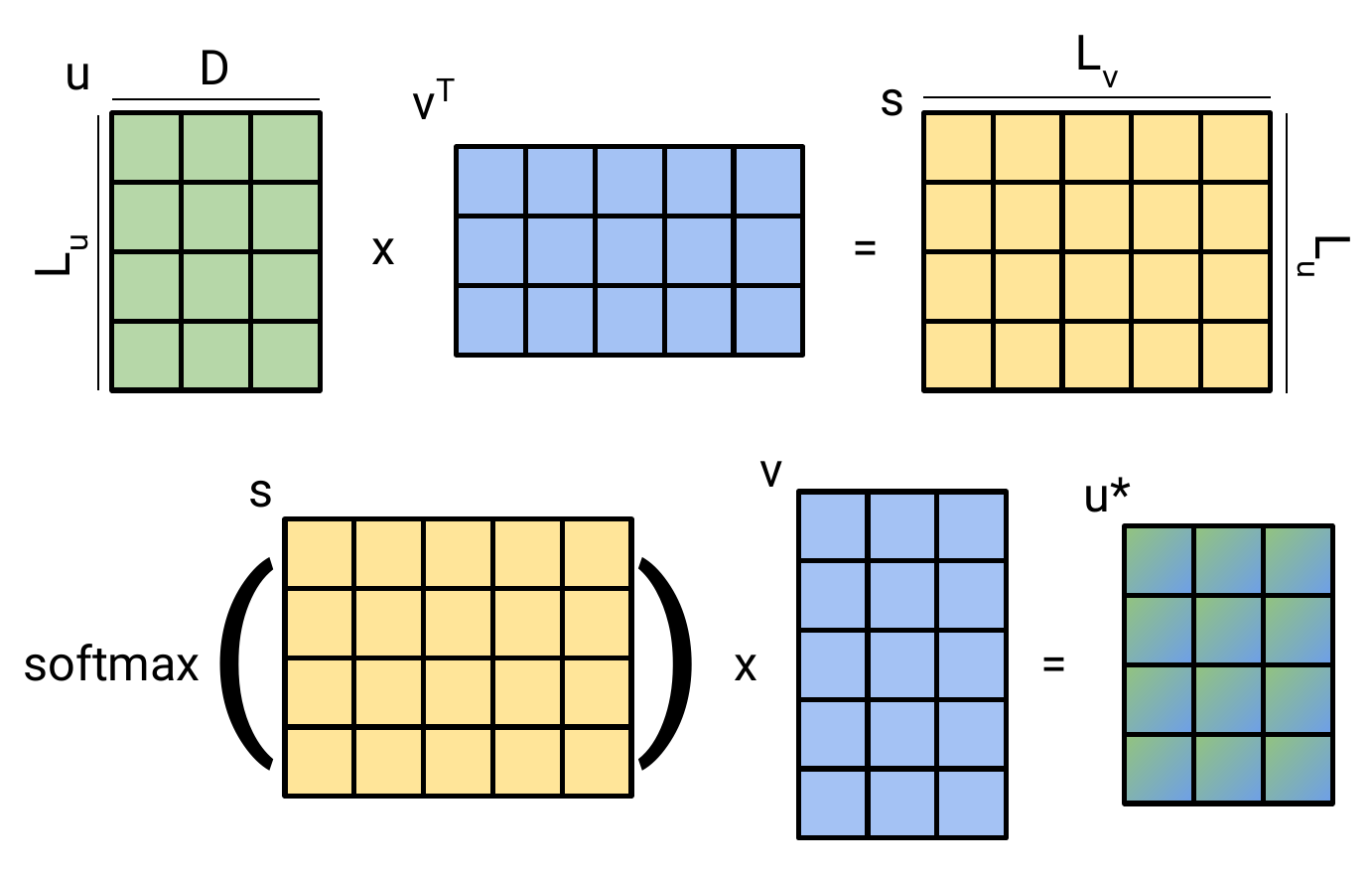}
    		\caption{Matrix alignment visualized, from $u$ and $v$ to $u^*$.}
    		\label{fig:0_matrix_alignment}
    	\end{figure*}

    \begin{equation}
    \label{eq:u_star}
    u^*=\text{softmax}(\frac{u\times v^T}{\sqrt{D}})\times v
    \end{equation}
        
    \begin{equation}
    \label{eq:v_star}
    v^*=\text{softmax}(\frac{v\times u^T}{\sqrt{D}})\times u
    \end{equation}
        
    where $u^T$ and $v^T$ are the transposed word sequences. Both operations are batch compatible if all positions at $u^*$ are set to $0$ where $u$ is $0$. The same is possible for $v^*$. This allows to ignore padding on matrix operations. 
        
    \item    \textbf{Step 4: Combination.} To extract richer information, the original and the aligned word sequences can be combined. For that purpose, the absolute difference $|u-v^*|$ and multiplication $|u\cdot v^*|$ are computed for sentence A and $|v^{*}-u|$ and $|v\cdot u^*|$ are calculated for sentence B. The absolute difference value amplifies high divergent values between original and aligned sequences, while the second operation amplifies high similar values. This same combination is also performed when merging both reasoning branches of the model, after step 5.
        
    \item    \textbf{Step 5: Multi-Head Self Attention block.} Multi-head self attention~\cite{vaswani2017attention} is a bi-directional block for sequence processing. Using the combined output of the aligned sequences, the self-attention block returns an interpreted sequence. The contents of the sequence are combined through a max pooling operator, returning the vectors $u_{max}$ $v_{max}$ of length $D$.
        
    The resulting $u_{max}$ and $v_{max}$ are concatenated with their respective $u_{cls}$ and $v_{cls}$ counterparts extracted after the embeddings are projected. The resulting vectors are named $x$ and $y$ for the left and right branch of the topology respectively. Then, both $x$ and $y$ are combined again using the absolute difference and the element-wise product, as explained in Step 4.
        
    \item   \textbf{Step 6: Feed-forward and output.} The combination of $x$ and $y$ is interpreted via a fully connected feed-forward layer with ReLU activation. Another feed-forward layer with three neurons (according to the three different categories) and softmax activation returns the outputs of the network.
        
    \item    \textbf{Regularization and optimization.} A dropout regularization is applied after the linear projection block, the self-attention block and the feed-forward layer. The dropout ratio is fine-tuned for each pre-trained transformer. The final linear layers are optimized with Adam~\cite{kingma2015adam}, following a cyclical learning rate~\cite{smith2017cyclical} with exponentially decaying upper bound.

\end{itemize}

\section{Experimental setup}
\label{sec:setup}

The experiments comprise the comparison of the SILT model with Sentence-BERT~\cite{reimers2019sentence} as a baseline against several state-of-the-art methods over two benchmark datasets: SICK~\cite{marelli2014sick} and MNLI~\cite{Williams_Nangia_Bowman_2018} (with validation over XNLI~\cite{Conneau_Rinott_Lample_Williams_Bowman_Schwenk_Stoyanov_2018}). Aside from the NLI performance of the SILT architecture, we assess the cross and inter-language capabilities as well. 

\subsection{Data}
\label{ssec:data}
We provide a summary of the two datasets used to evaluate the SILT architecture in Table \ref{tab:datasum}. SICK, which is a light dataset with simple sentences, where the low size is a challenge due to its tendency to overfit, and MNLI+XNLI, a state-of-the-art NLI task, with almost half a million sentence pairs with complex syntax and a variety of genres and languages.

\begin{table}[]
\centering
\resizebox{\textwidth}{!}{%
\begin{tabular}{@{}c|cccc|cccc|cccc@{}}
\toprule
\textbf{}      & \multicolumn{4}{c|}{\textbf{Contradiction}} & \multicolumn{4}{c|}{\textbf{Entailment}} & \multicolumn{4}{c|}{\textbf{Neutral}} \\ \cmidrule(l){2-13} 
\multirow{2}{*}{\textbf{Subset}} &
  \multicolumn{1}{c|}{\textbf{SICK}} &
  \multicolumn{3}{c|}{\textbf{SICK-ES}} &
  \multicolumn{1}{c|}{\textbf{SICK}} &
  \multicolumn{3}{c|}{\textbf{SICK-ES}} &
  \multicolumn{1}{c|}{\textbf{SICK}} &
  \multicolumn{3}{c|}{\textbf{SICK-ES}} \\ \cmidrule(l){2-13} 
 &
  \multicolumn{1}{c|}{\textbf{En-En}} &
  \textbf{Es-En} &
  \textbf{En-Es} &
  \textbf{Es-Es} &
  \multicolumn{1}{c|}{\textbf{En-En}} &
  \textbf{Es-En} &
  \textbf{En-Es} &
  \textbf{Es-Es} &
  \multicolumn{1}{c|}{\textbf{En-En}} &
  \textbf{Es-En} &
  \textbf{En-Es} &
  \textbf{Es-Es} \\ \midrule\midrule
\textbf{Train} & 641       & 641       & 641      & 641      & 1274     & 1274     & 1274     & 1274    & 2524    & 2524    & 2524    & 2524    \\
\textbf{Valid} & 71        & 71        & 71       & 71       & 143      & 143      & 143      & 143     & 281     & 281     & 281     & 281     \\
\textbf{Test}  & 712       & 712       & 712      & 712      & 1404     & 1404     & 1404     & 1404    & 2790    & 2790    & 2790    & 2790    \\ \midrule
 &
  \multicolumn{1}{c|}{\textbf{MNLI}} &
  \multicolumn{3}{c|}{\textbf{XNLI}} &
  \multicolumn{1}{c|}{\textbf{MNLI}} &
  \multicolumn{3}{c|}{\textbf{XNLI}} &
  \multicolumn{1}{c|}{\textbf{MNLI}} &
  \multicolumn{3}{c|}{\textbf{XNLI}} \\ \cmidrule(l){2-13} 
\textbf{} &
  \multicolumn{1}{c|}{\textbf{En-En}} &
  \textbf{En-Es} &
  \textbf{Es-En} &
  \textbf{Es-Es} &
  \multicolumn{1}{c|}{\textbf{En-En}} &
  \textbf{En-Es} &
  \textbf{Es-En} &
  \textbf{Es-Es} &
  \multicolumn{1}{c|}{\textbf{En-En}} &
  \textbf{En-Es} &
  \textbf{Es-En} &
  \textbf{Es-Es} \\ \midrule\midrule
\textbf{Train} & 130903    & 130903    & 130903   & 130903   & 130899   & 130899   & 130899   & 130899  & 130900  & 130900  & 130900  & 130900  \\
\textbf{Valid} & 830       & 830       & 830      & 830      & 830      & 830      & 830      & 830     & 830     & 830     & 830     & 830     \\
\textbf{Test}  & 1670      & 1670      & 1670     & 1670     & 1670     & 1670     & 1670     & 1670    & 1670    & 1670    & 1670    & 1670    \\ \bottomrule
\end{tabular}%
}
\caption{Summarization of labels within train, test, and validation sets of the proposed datasets. The SICK-ES column represents the machine translated version of SICK, which in the case of the test set is manually curated. }
\label{tab:datasum}
\end{table}

\textbf{SICK and SICK-ES:} This dataset is a benchmark for bidirectional textual entailment and similarity. It consists of 10k English sentences, labeled with the entailment and level of similarity, along other metadata. Sentences in this dataset are usually short ($<100$ characters) and are straightforward in meaning. Since this dataset only includes English text, we propose and make publicly available a testing set called SICK-ES, where a subset of sentences from SICK are machine translated into Spanish and later manually curated to better represent Spanish sentence composition. This enables to extend this dataset to a multilingual scenario. In order to evaluate cross-language capabilities, Spanish sentences are also compared with their English sentence counterpart, and vice-versa. The SICK training set was also machine translated in order to fit the neural network with cross-language capabilities. 

\textbf{MNLI and XNLI:} MNLI is a large dataset of sentence pairs from different genres, consisting of 433k annotated examples. Accurate classification requires to understand long sentences (sometimes $>500$ characters) and different literature genres, without context (if mismatched). This is a state-of-the-art benchmark featured in GLUE~\cite{wang2019glue} tasks. XNLI, on the other hand, is a replacement for the MNLI testing set, featuring paired multi-language sentences for the evaluation of MNLI. Usual practices require to machine translate MNLI to later evaluate on XNLI, which in this case is already provided as XNLI-MT. As it happened on SICK, only the testing set (XNLI) is curated, while the MNLI machine translation is not.

\subsection{Metrics}
\label{ssec:metrics}
As a classification problem, we account for three different labels: entailment, contradiction and neutral. From the predictions of these labels, we can extract metrics such as \textit{precision}, \textit{recall} and \textit{F1-score}, as well as \textit{accuracy}. During the experiment, F1-score and accuracy are reported as we specifically require to detect how well classified are underrepresented classes (for example, while SICK has a total of 3k examples of entailment, it only contains 1.4k examples of contradictions). 

\subsection{Base transformers}
\label{ssec:base_transformer}
Our model relies on a frozen pre-trained transformer to extract sentence features. These transformers have been selected using the Huggingface library~\cite{Wolf_Debut_Sanh_Chaumond_Delangue_Moi_Cistac_Rault_Louf_Funtowicz_2019}, which has three out-of-the-box transformers with multi-lingual tokenization and pre-training with different levels of complexity. The selected models require to be multilingual (cross-lingual feature extraction): 

\textbf{BERT-base}~\cite{kentonbert}: BERT popularized the transformer as a way to understand text. It is pre-trained with a large text corpus and out-of-the-box is capable to perform well on GLUE tasks, and with some added fine-tuning it can perform even better. It is a hallmark of natural language processing and several variants have been proposed. When pre-trained with a multilingual corpus it is able to recognize sentences from multiple languages, which is a requirement for cross-lingual tasks. The multi-lingual version is used in this article when we refer to BERT-base.

\textbf{DistilBERT}~\cite{Sanh_Debut_Chaumond_Wolf_2019}: Knowledge distillation is a powerful resource to optimize networks into a lower parameter space. DistilBERT is a distilled version of BERT than holds 95\% of its original capabilities with half the transformer layers (from 12 in BERT-base to 6 in DistilBERT). As such more information is compressed in the intermediate layers, and quicker computations can be made. These compressed representations may be easier to identify by the network than the original BERT embeddings.

\textbf{XLM-RoBERTa-base (XLM-R)}~\cite{conneau2020unsupervised}: XLM-R significantly outperforms multilingual BERT on multilingual tasks, including XNLI. This model uses an XLM~\cite{Lample_Conneau_2019} architecture enhanced with slight performance changes, such as the text corpus size (1 order of magnitude higher). This increase in performance may mean that XLM-R embeddings are much more representative of the tokenized sentences. 

\subsection{Hyper-parameters}
\label{ssec:params}
Hyper-parameters have been tuned using the provided validation subset of each benchmark, aiming to minimize loss. Inner layers of the SILT topology have a set embedding dimension of $768$ with $8$ attention heads and a feed-forward layer dimension of $768$ with ReLU activation. In order to avoid overfitting, a dropout layer is used with adjusted value for each transformer and data. When training on SICK, it was shown that the SILT architecture is sensitive to the dropout rate, so it was tuned independently for each model, with $d=0.4$ for distilBERT, $d=0.3$ for BERT and $d=0.35$ for XLM-R. Dropout on XNLI has not been tuned, being set to $d=0.1$, as with enough data this tuning is not needed.

Optimization is done through Adam, with $\beta_1=0.9$, $\beta_2=0.9999$ and $\epsilon=1\cdot10^{-7}$. The learning rate cycles with exponential decay starting from an initial learning rate of $\alpha_0=3\cdot10^{-6}$ and a maximum of $\alpha_{max}=3\cdot10^{-3}$.

\subsection{Baselines}
\label{ssec:baseline}
As a custom baseline, we use Sentence-BERT (S-BERT) to evaluate if the sentence pair alignment is effective at capturing relations between sentences. This design is powerful on sentence-pair tasks, as proved on Semantic Textual Similarity (STS) tasks. However, the performance when the transformer is frozen is unknown. Thus, we focus on assessing if the transformer with frozen weights is powerful enough to produce accurate results by training a Linear-Softmax layer. Aside from S-BERT, both MNLI and SICK come with a benchmark of results, as well as BERT, DistilBERT and XLM-R, which will be discussed as part of our comparison.


     
     
     
     

\begin{table}[]
\centering
\resizebox{\textwidth}{!}{%
\begin{tabular}{@{}lllcccc@{}}
\toprule
\multicolumn{1}{c}{\textbf{Dataset}} &
  \multicolumn{1}{c}{\textbf{Topology}} &
  \multicolumn{1}{c}{\textbf{Base transformer}} &
  \textbf{Accuracy} &
  \textbf{F1-Score} &
  \textbf{\#Param.} &
  \textbf{\#Train.} \\ \midrule\midrule
\multirow{6}{*}{\textbf{SICK}} & \multirow{3}{*}{\textbf{Sentence-BERT}} & \multicolumn{1}{l|}{\textbf{BERT}}       & 49.64\%          & 49.36\%          & $1.78\cdot10^8$ & $1.2\cdot10^5$ \\
                               &                                  & \multicolumn{1}{l|}{\textbf{XLM-R}}      & 57.46\%          & 45.21\%          & $2.78\cdot10^8$ & $1.2\cdot10^5$ \\
                               &                                  & \multicolumn{1}{l|}{\textbf{distilBERT}} & 57.92\%          & 37.56\%          & $1.34\cdot10^8$&  $6.4\cdot10^4$  \\ \cmidrule(l){2-7} 
                               & \multirow{3}{*}{\textbf{SILT}}    & \multicolumn{1}{l|}{\textbf{BERT}}       & 79.05\%          & 75.73\%          & $1.94\cdot10^8$ & $1.7\cdot10^7$  \\
                               &                                  & \multicolumn{1}{l|}{\textbf{XLM-R}}      & \textbf{82.47\%} & 81.52\%          & $2.95\cdot10^8$ & $1.7\cdot10^7$  \\
                               &                                  & \multicolumn{1}{l|}{\textbf{distilBERT}} & 82.35\%          & \textbf{81.64\%} & $1.48\cdot10^8$ & $1.3\cdot10^7$  \\ \midrule
\multirow{6}{*}{\textbf{XNLI}} & \multirow{3}{*}{\textbf{Sentence-BERT}} & \multicolumn{1}{l|}{\textbf{BERT}}       & 51.86\%          & 47.78\%          & $1.78\cdot10^8$  & $1.2\cdot10^5$  \\
                               &                                  & \multicolumn{1}{l|}{\textbf{XLM-R}}      & 57.46\%          & 45.21\%          & $2.78\cdot10^8$ & $1.2\cdot10^5$  \\
                               &                                  & \multicolumn{1}{l|}{\textbf{distilBERT}} & 53.56\%          & 51.43\%          & $1.34\cdot10^8$ & $6.4\cdot10^4$   \\ \cmidrule(l){2-7} 
                               & \multirow{3}{*}{\textbf{SILT}}    & \multicolumn{1}{l|}{\textbf{BERT}}       & 72.73\%          & 72.65\%          & $1.94\cdot10^8$ & $1.7\cdot10^7$  \\
                               &                                  & \multicolumn{1}{l|}{\textbf{XLM-R}}      & \textbf{79.29\%} & \textbf{79.17\%} & $2.95\cdot10^8$ & $1.7\cdot10^7$  \\
                               &                                  & \multicolumn{1}{l|}{\textbf{distilBERT}} & 73.54\%          & 73.48\%          & $1.48\cdot10^8$ & $1.3\cdot10^7$  \\ \bottomrule
\end{tabular}%
}
\caption{Results obtained in the test set of SICK and XNLI datasets with SILT and the baseline model SBERT, reporting accuracy, f1-score and network size (total number of parameters and trainable parameters) for each base transformer.}
\label{tab:0_summary}
\end{table}

\section{Results and discussion}
\label{sec:results}

The experiments to assess the SILT architecture have been divided into three groups:
\begin{enumerate}
    \item Comparison against S-BERT baseline. Our motivation is to evaluate if the frozen networks have enough discrimination power to separate clasess by itself. If this is not the case, the addition of a reasoning head (SILT) is justified.
    \item Comparison against state-of-the-art models. This allows us to check where our model stands in terms of overall performance, and how the inter-linugal capabilities of SILT compare against academic standards.
    \item SILT evaluation. An exploration on the performance controlling by different characteristics: Sentence similarity, literary genre, etc. This is aimed toward analyzing the strengths and weak points of SILT.
\end{enumerate}

\subsection{Baseline study}
The first experiment evaluates the performance of the SILT architecture in comparison with a baseline model using the Sentence-BERT architecture. The evaluation is performed in both SICK and XNLI datasets for each pre-trained transformer considered: BERT, XLM-R and distilBERT. Overall accuracy and F1-scores of this experiment are reported in Table~\ref{tab:0_summary}, together with the total number of parameters and number of trainable parameters.

As can be seen in this table, Sentence-BERT is vastly inferior to the SILT topology, resulting in a drop of at least 20\% in accuracy and F1-score independently of the base transformer used. This confirms our previous assumption that architectures with frozen transformers such as S-BERT do not work properly in isolation, requiring additional reasoning layers to adequately undertake a specific task. We also observe that, on average, accuracy and F1-score are fairly similar across models, so any imbalance in the data has proven to be irrelevant for both topologies in this experiment.

Regarding the size of the architecture, the number of trainable parameters for Sentence-BERT is lower than those for SILT (see \textit{\#Train} column), a factor which strongly influences the drop in accuracy. The SILT topology adds a head with 13-17 million parameters, which achieves higher performance. This means the embeddings of the transformer could be cached, bypassing at least 130 million parameters in training computation.

\subsection{Performance evaluation against state-of-the-art architectures for NLI}
\begin{table}[]
\centering
\resizebox{\textwidth}{!}{%
\begin{tabular}{@{}clcccc@{}}
\toprule
\textbf{} &
  \multicolumn{1}{c}{\textbf{Method}} &
  \textbf{\begin{tabular}[c]{@{}c@{}}\#Trainable\\ parameters\end{tabular}} &
  \textbf{En-En} &
  \textbf{En-Es} &
  \textbf{Es-Es} \\ \midrule\midrule
\multirow{5}{*}{\textbf{\begin{tabular}[c]{@{}c@{}}XNLI benchmark\\ methods\end{tabular}}} &
  \multicolumn{1}{l|}{\textbf{BiLSTM-last \cite{Conneau_Rinott_Lample_Williams_Bowman_Schwenk_Stoyanov_2018}}} &
  \multicolumn{1}{c|}{$3\cdot10^6$} &
  71.0\% &
  - &
  67.0\% \\
 &
  \multicolumn{1}{l|}{\textbf{BiLSTM-max \cite{Conneau_Rinott_Lample_Williams_Bowman_Schwenk_Stoyanov_2018}}} &
  \multicolumn{1}{c|}{$3\cdot10^6$} &
  \textbf{73.7\%} &
  - &
  \textbf{68.8\%} \\
 &
  \multicolumn{1}{l|}{\textbf{X-BiLSTM-Last \cite{Conneau_Rinott_Lample_Williams_Bowman_Schwenk_Stoyanov_2018}}} &
  \multicolumn{1}{c|}{$3\cdot10^6$} &
  71.0\% &
  - &
  67.8\% \\
 &
  \multicolumn{1}{l|}{\textbf{X-BiLSTM-max \cite{Conneau_Rinott_Lample_Williams_Bowman_Schwenk_Stoyanov_2018}}} &
  \multicolumn{1}{c|}{$3\cdot10^6$} &
  \textbf{73.7\%} &
  - &
  68.7\% \\
 &
  \multicolumn{1}{l|}{\textbf{X-CBOW \cite{Conneau_Rinott_Lample_Williams_Bowman_Schwenk_Stoyanov_2018}}} &
  \multicolumn{1}{c|}{$1.5\cdot10^8$} &
  64.5\% &
  - &
  60.7\% \\ \midrule
\multirow{4}{*}{\textbf{\begin{tabular}[c]{@{}c@{}}Current state-of-the-art\\ cross-lingual NLI\\ approaches\end{tabular}}} &
  \multicolumn{1}{l|}{\textbf{FILTER \cite{Fang_Wang_Gan_Sun_Liu_2020}}} &
  \multicolumn{1}{c|}{$2.78\cdot10^8$} &
  89.5\% &
  - &
  \textbf{86.6\%} \\
 &
  \multicolumn{1}{l|}{\textbf{XLM-R R4F \cite{Aghajanyan_Shrivastava_Gupta_Goyal_Zettlemoyer_Gupta_2020}}} &
  \multicolumn{1}{c|}{$5.59\cdot10^8$} &
  \textbf{89.6\%} &
  - &
  85.2\% \\
 &
  \multicolumn{1}{l|}{\textbf{XLM-R Base \cite{Aghajanyan_Shrivastava_Gupta_Goyal_Zettlemoyer_Gupta_2020}}} &
  \multicolumn{1}{c|}{$2.78\cdot10^8$} &
  85.8\% &
  - &
  80.7\% \\
 &
  \multicolumn{1}{l|}{\textbf{Unicoder \cite{huang2019unicoder}}} &
  \multicolumn{1}{c|}{$1.42\cdot10^8$} &
  85.6\% &
  - &
  81.1\%  \\ \midrule 
\multirow{3}{*}{\textbf{SILT}} &
  \multicolumn{1}{l|}{\textbf{SILT+XLM-R}} &
  \multicolumn{1}{c|}{$1.7\cdot10^7$} &
  \textbf{81.0\%} &
  \textbf{79.1\%} &
  \textbf{77.9\%} \\
 &
  \multicolumn{1}{l|}{\textbf{SILT+BERT}} &
  \multicolumn{1}{c|}{$1.7\cdot10^7$} &
  74.6\% &
  72.0\% &
  72.3\% \\
 &
  \multicolumn{1}{l|}{\textbf{SILT+distilBERT}} &
  \multicolumn{1}{c|}{\textbf{$1.3\cdot10^7$}} &
  75.5\% &
  73.2\% &
  72.3\% \\ \bottomrule
\end{tabular}%
}
\caption{Accuracy reported for several state-of-the-art published methods on the XNLI benchmark. The source of the reported measurements is cited along with the method. The first group reports the benchmark of methods that were used in the original publication of XNLI~\cite{Conneau_Rinott_Lample_Williams_Bowman_Schwenk_Stoyanov_2018}. The second groups summaries current state-of-the-art approaches on XNLI. The third group reports the results of the proposed architecture SILT along different base transformers.}
\label{tab:6_comparative_to_sota_XNLI}
\end{table}

Comparisons with state-of-the-art models are offered in Tables \ref{tab:6_comparative_to_sota_XNLI} and \ref{tab:7_comparative_to_sota_SICK} for XNLI and SICK respectively. For the XNLI comparison, we find three blocks in the table, with the original XNLI benchmark, the current best cross-lingual approaches to XNLI and the SILT approach with different base transformers. First, it is important to note that English+Spanish (among other inter-lingual pairs) are rarely, if ever, discussed in the literature, which focuses mainly on English+English pairs or Spanish+Spanish pairs, always using matched language pairs at the input level. From the table, it can be seen that every accuracy measurement is always better for English+English pairs, which shows clear bias towards the native language of the model (commonly english) inherent to models fine-tuned with MNLI+XNLI.

\begin{table}[htpb]
\centering
\resizebox{0.7\textwidth}{!}{%
\begin{tabular}{@{}lcccc@{}}
\toprule
\multicolumn{1}{c}{\multirow{2}{*}{\textbf{Method}}} &
  \multirow{2}{*}{\textbf{\begin{tabular}[c]{@{}c@{}}\#Trainable\\ parameters\end{tabular}}} &
  \multicolumn{3}{c}{\textbf{Accuracy on SICK dataset}} \\ \cmidrule(l){3-5} 
\multicolumn{1}{c}{}                                             &                & \textbf{En-En} & \textbf{En-Es} & \textbf{Es-Es} \\ \midrule\midrule
\textbf{GenSen \cite{subramanian2018learning}}  & $5\cdot10^7$   & 87,8\%         & -              & -              \\
\textbf{InferSent \cite{conneau2017supervised}} & $3.8\cdot10^7$ & 86,5\%         & -              & -              \\ \midrule
\textbf{SILT+XLM-R}                             & $1.7\cdot10^7$ & 83.3\%         & 81.3\%         & 84.0\%         \\
\textbf{SILT+BERT}                              & $1.7\cdot10^7$ & 82.3\%         & 75.2\%         & 83.5\%         \\
\textbf{SILT+distilBERT}                        & $1.3\cdot10^7$ & 84.8\%         & 79.9\%         & 84.7\%         \\ \bottomrule
\end{tabular}%
}
\caption{Accuracy reported for several state-of-the-art published methods on the SICK benchmark. The source of the reported measurements is cited along with the method.}
\label{tab:7_comparative_to_sota_SICK}
\end{table}

Common criticism of cross-lingual systems include the reduction against mono-lingual models. While SILT topologies clearly outperform the provided XNLI benchmark results, current state of the art methods are more robust both in English and Spanish pairs. While these values exceed the results obtained with SILT, this comes at a high training cost. For instance, XLM-R Base uses 278M parameters, and is very time-consuming to fine-tune (more to pre-train) on GPUs. SILT with a frozen XLM-R and a number of trainable parameters one order of magnitude lower, achieves similar results across matched and mixed language pairs, thus avoiding overfitting to a specific language while keeping inter-lingual capabilities.

Table~\ref{tab:7_comparative_to_sota_SICK} shows the results of SILT in the SICK dataset along with two state-of-the-art models in this dataset. As can be seen, neither GenSen nor InferSent have cross-lingual capabilities and have been designed only for English.The results point that even when learning multiple languages, SILT remains competitive with a slight drop in accuracy ranging from 1\% to 2\% compared with the top methods found in this category. This is done by training less parameters, while adding inter-lingual capabilities. Despite the model inter-linguality, we observe that for SICK a small drop in accuracy happens from matched language pairs, to mixed language pairs. 


\subsection{Evaluation of SILT results}

The following tables break down accuracy and F1-score by a variety of criteria. First, Table \ref{tab:1_results_per_label} shows the specific accuracy by true label. First, results on SICK are clearly biased towards predicting neutral statements, as opposed to contradictions, with entailments in the middle. Specifically when using BERT and distilBERT as a base, the increase in accuracy of Neutral statements is above 10\%. The opposite happens on XNLI, where results are clearly biased toward predicting contradictions, despite being a balanced dataset. On SICK and XNLI, the XLM-R base is the least biased of the three base transformers.





\begin{table}[htpb]
\centering
\resizebox{\textwidth}{!}{%
\begin{tabularx}{1.3\textwidth}{@{}llYYY@{}}
\toprule
\multirow{2}{*}{\textbf{Dataset}} & \multirow{2}{*}{\textbf{Base transformer}} & \multicolumn{3}{c}{\textbf{NLI result by label}} \\ \cmidrule(l){3-5} 
                               &                                          & \textbf{Contradiction} & \textbf{Entailment} & \textbf{Neutral} \\ \midrule\midrule
\multirow{3}{*}{\textbf{SICK}}    & \multicolumn{1}{l|}{\textbf{BERT}}         & 57.30\%     & 69.68\%     & \textbf{89.32\%}     \\
                               & \multicolumn{1}{l|}{\textbf{XLM-R}}      & \textbf{79.00\%}       & \textbf{80.61\%}    & 84.28\%          \\
                               & \multicolumn{1}{l|}{\textbf{distilBERT}} & 76.86\%                & 77.23\%             & 86.33\%          \\ \midrule
\multirow{3}{*}{\textbf{XNLI}} & \multicolumn{1}{l|}{\textbf{BERT}}       & 81.68\%                & 69.24\%             & 67.28\%          \\
                               & \multicolumn{1}{l|}{\textbf{XLM-R}}      & \textbf{84.87\%}       & \textbf{82.19\%}    & \textbf{70.81\%} \\
                               & \multicolumn{1}{l|}{\textbf{distilBERT}} & 78.85\%                & 73.58\%             & 68.19\%          \\ \bottomrule
\end{tabularx}%
}
\caption{Results in terms of accuracy obtained with the SILT topology in the SICK and XNLI datasets grouped by label: contradiction, entailment, neutral.}
\label{tab:1_results_per_label}
\end{table}

\begin{figure}[htpb]  
    \centering
    \includegraphics[width=0.8\textwidth]{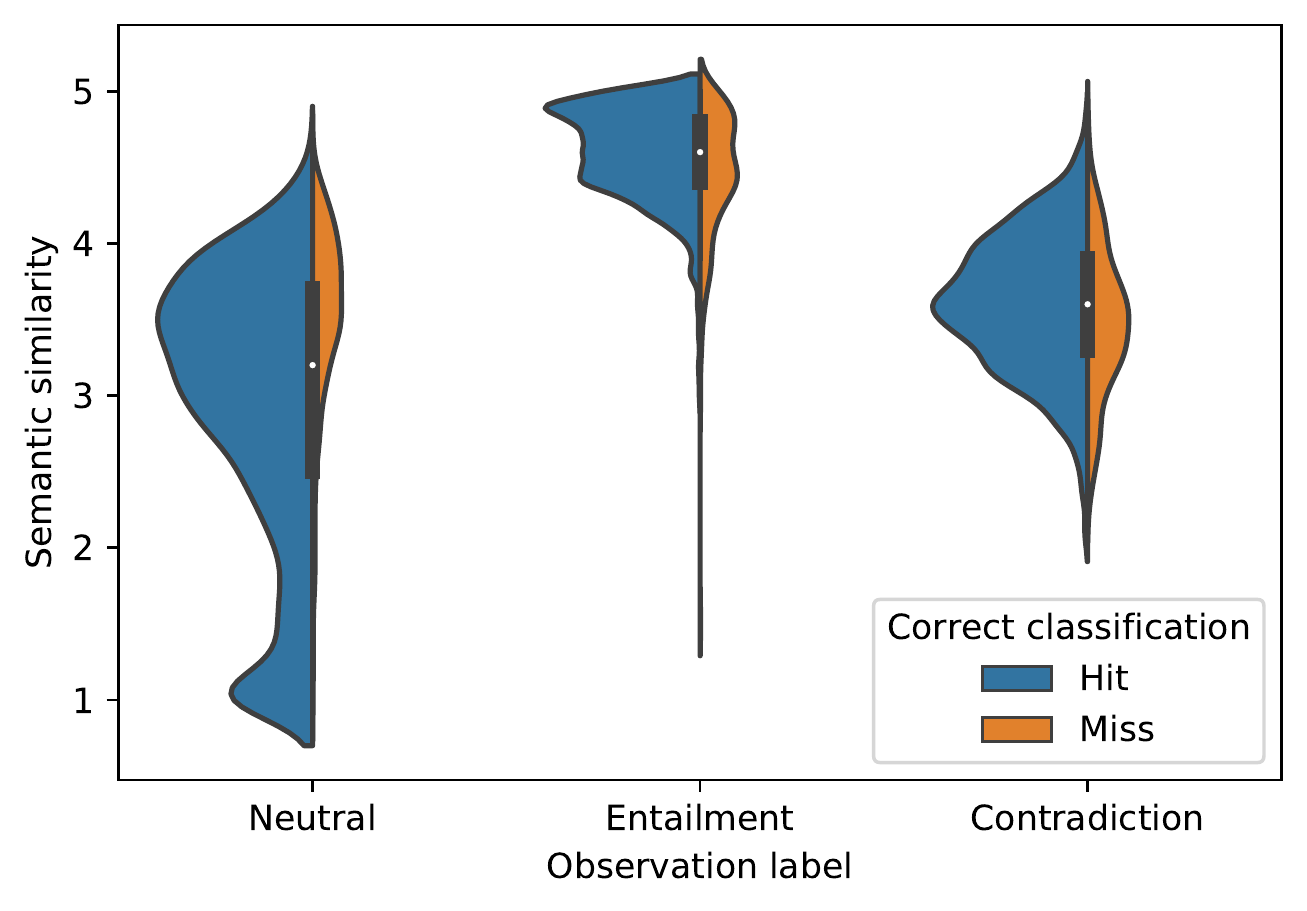}
    \caption{Distribution of hits and misses by label against their semantic similarity, generated by the SILT topology with XLM-R base transformer.}
    \label{fig:2_accuracy_semantic}
\end{figure}

Further explanation of this bias is explored in Table \ref{tab:4_results_by_semantics_sick} where results are grouped by semantic similarity. Here we find a large mismatch between accuracy and F1-score, unseen in previous tables, for very low similarities, accuracy is extremely high and F1-score is a the worst-case scenario. This happens because the relatedness score of the dataset is itself biased, as illustrated on Figure \ref{fig:2_accuracy_semantic}. 




\begin{table}[htpb]
\centering
\resizebox{\textwidth}{!}{%
\begin{tabularx}{1.3\textwidth}{@{}llYYYY@{}}
\toprule
\multirow{2}{*}{\textbf{Metric}} &
  \multirow{2}{*}{\textbf{Base transformer}} &
  \multicolumn{4}{c}{\textbf{Sentece relatedness score range}} \\ \cmidrule(l){3-6} 
 &
   &
  \multicolumn{1}{c}{\textbf{(1, 2{]}}} &
  \multicolumn{1}{c}{\textbf{(2, 3{]}}} &
  \multicolumn{1}{c}{\textbf{(3, 4{]}}} &
  \multicolumn{1}{c}{\textbf{(4, 5{]}}} \\ \midrule
\multirow{3}{*}{\textbf{Accuracy}} & \multicolumn{1}{l|}{\textbf{BERT}}       & 98.08\% & 90.65\% & 76.22\% & 71.18\% \\
                                   & \multicolumn{1}{l|}{\textbf{XLM-R}}      & 95.17\% & 89.37\% & 78.46\% & 79.88\% \\
                                   & \multicolumn{1}{l|}{\textbf{distilBERT}} & 96.25\% & 90.72\% & 79.03\% & 78.09\% \\ \midrule
\multirow{3}{*}{\textbf{F1-Score}} & \multicolumn{1}{l|}{\textbf{BERT}}       & 33.01\% & 52.56\% & 58.45\% & 64.34\% \\
                                   & \multicolumn{1}{l|}{\textbf{XLM-R}}      & 32.51\% & 57.38\% & 63.84\% & 71.99\% \\
                                   & \multicolumn{1}{l|}{\textbf{distilBERT}} & 32.70\% & 59.27\% & 64.16\% & 73.40\% \\ \bottomrule
\end{tabularx}%
}
\caption{Evaluation of SILT performance in the SICK dataset depending on the relatedness of each pair of sentences, as defined by the authors of SICK~\cite{marelli2014sick}. The results show the accuracy and f1-score values for each base transformer tested. Sentences with a neutral relation are distributed across the whole range of the relatedness score, but in large part present when the score is lower than 4. Pairs of sentences with contradiction relation can be primarily found when the relatedness core ranges from 3 to 4, while entailment mainly occurs when the score is higher than 4.}
\label{tab:4_results_by_semantics_sick}
\end{table}

We find that the distribution of labels has a strong relationship with semantic similarity, meaning that the model is detecting some measure of relatedness and discriminating accordingly. For example for similarities between 1 and 1.5, the label is frequently Neutral and the model reacts accordingly. Similarly, when an entailment relation has less than relatedness 4, the model starts missing more. This offers a sensible explanation on why biases arise when breaking down results by label or relatedness: the model is discovering a relatedness score without explicit information. 

On Table \ref{tab:2_results_per_language} we find a breakdown of results by language, including inter-lingual sentence pairs. On the SICK dataset cross-lingual pairs are usually worse classified than language matched sentence pairs, with drops of around 7\% (for BERT), 5\% (for distilBERT) and 3\% (for XLM-R). On the other hand, the XNLI evaluation set is worse at classifying Spanish sentence pairs, although oscillations in Accuracy and F1-Score is lower than 2\%.


      
 
         
         



\begin{table}[htpb]
\centering
\resizebox{\textwidth}{!}{%
\begin{tabularx}{1.3\textwidth}{@{}lllYYY@{}}

\toprule
\multicolumn{1}{c}{\multirow{2}{*}{\textbf{Metric}}} &
  \multicolumn{1}{c}{\multirow{2}{*}{\textbf{Dataset}}} &
  \multicolumn{1}{c}{\multirow{2}{*}{\textbf{Base transformer}}} &
  \multicolumn{3}{c}{\textbf{Language}} \\ \cmidrule(l){4-6} 
\multicolumn{1}{c}{}               & \multicolumn{1}{c}{}           & \multicolumn{1}{c}{}                     & \textbf{En-En}   & \textbf{En-Es}   & \textbf{Es-Es}   \\ \midrule\midrule
\multirow{6}{*}{\textbf{Accuracy}} & \multirow{3}{*}{\textbf{SICK}} & \multicolumn{1}{l|}{\textbf{BERT}}       & 82.27\%          & 75.24\%          & 83.45\%          \\
                                   &                                & \multicolumn{1}{l|}{\textbf{XLM-R}}      & 83.33\%          & \textbf{81.27\%} & 84.00\%          \\
                                   &                                & \multicolumn{1}{l|}{\textbf{distilBERT}} & \textbf{84.77\%} & 79.94\%          & \textbf{84.73\%} \\ \cmidrule(l){2-6} 
                                   & \multirow{3}{*}{\textbf{XNLI}} & \multicolumn{1}{l|}{\textbf{BERT}}       & 74.57\%          & 72.04\%          & 72.28\%          \\
                                   &                                & \multicolumn{1}{l|}{\textbf{XLM-R}}      & \textbf{81.04\%} & \textbf{79.11\%} & \textbf{77.88\%} \\
                                   &                                & \multicolumn{1}{l|}{\textbf{distilBERT}} & 75.47\%          & 73.21\%          & 72.26\%          \\ \midrule
\multirow{6}{*}{\textbf{F1-Score}} & \multirow{3}{*}{\textbf{SICK}} & \multicolumn{1}{l|}{\textbf{BERT}}       & 81.78\%          & 68.43\%          & 81.83\%          \\
                                   &                                & \multicolumn{1}{l|}{\textbf{XLM-R}}      & 82.75\%          & \textbf{79.84\%} & 83.36\%          \\
                                   &                                & \multicolumn{1}{l|}{\textbf{distilBERT}} & \textbf{84.61\%} & 78.76\%          & \textbf{84.22\%} \\ \cmidrule(l){2-6} 
                                   & \multirow{3}{*}{\textbf{XNLI}} & \multicolumn{1}{l|}{\textbf{BERT}}       & 74.25\%          & 71.95\%          & 72.20\%          \\
                                   &                                & \multicolumn{1}{l|}{\textbf{XLM-R}}      & \textbf{80.90\%} & \textbf{78.99\%} & \textbf{77.79\%} \\
                                   &                                & \multicolumn{1}{l|}{\textbf{distilBERT}} & 75.24\%          & 73.20\%          & 72.20\%          \\ \bottomrule
\end{tabularx}%
}
\caption{Results obtained with SILT for language inference in terms of accuracy and f1-score according to the language of the pair of sentences in the SICK and XNLI testing sets. \textit{En-En} means that sentences are paired in English, while \textit{Es-Es} are paired in Spanish. Lastly, \textit{En-Es} is used for cross-lingual inference, where a sentence in Spanish and another in English.}
\label{tab:2_results_per_language}
\end{table}

Further analysis is conducted on Table \ref{tab:3_results_by_genre_xnli} where accuracy is measured by genre. Each genre is distinct and the model presents further biases. On XLM-R, Government language has the highest accuracy (83.5\%), while the hardest genre is verbatim (72.41\%). The rest of genres lie in-between that range, and there is no clear bias towards informal registries (face to face or travel), formal registries (government) or even fictional text, although the biases prevail from a base transformer to others, where Verbatim is always the worst classified and Government the best.

\begin{table}[htpb]
\centering
\resizebox{\textwidth}{!}{%
\begin{tabularx}{1.4\textwidth}{@{}llYYY@{}}
\toprule
\multirow{2}{*}{\textbf{Metric}} &
  \multirow{2}{*}{\textbf{\begin{tabular}[c]{@{}c@{}}Genre of text\\ for XNLI\end{tabular}}} &
  \multicolumn{3}{c}{\textbf{Base transformer}} \\ \cmidrule(l){3-5} 
                                    &                                          & \textbf{BERT} & \textbf{XLM-R} & \textbf{distilBERT} \\ \midrule\midrule
\multirow{10}{*}{\textbf{Accuracy}} & \multicolumn{1}{c|}{\textbf{facetoface}} & 73.75\%       & 79.99\%        & 76.60\%             \\
                                    & \multicolumn{1}{c|}{\textbf{fiction}}    & 72.95\%       & 78.89\%        & 74.60\%             \\
                                    & \multicolumn{1}{c|}{\textbf{government}} & 76.90\%       & 83.53\%        & 77.05\%             \\
                                    & \multicolumn{1}{c|}{\textbf{letters}}    & 73.70\%       & 82.73\%        & 76.70\%             \\
                                    & \multicolumn{1}{c|}{\textbf{nineeleven}} & 76.10\%       & 82.58\%        & 76.75\%             \\
                                    & \multicolumn{1}{c|}{\textbf{oup}}        & 70.76\%       & 76.35\%        & 70.66\%             \\
                                    & \multicolumn{1}{c|}{\textbf{slate}}      & 69.01\%       & 75.55\%        & 68.71\%             \\
                                    & \multicolumn{1}{c|}{\textbf{telephone}}  & 70.71\%       & 77.79\%        & 72.01\%             \\
                                    & \multicolumn{1}{c|}{\textbf{travel}}     & 75.20\%       & 83.03\%        & 73.95\%             \\
                                    & \multicolumn{1}{c|}{\textbf{verbatim}}   & 68.21\%       & 72.41\%        & 68.36\%             \\ \midrule
\multirow{10}{*}{\textbf{F1-Score}} & \multicolumn{1}{c|}{\textbf{facetoface}} & 73.64\%       & 79.87\%        & 76.50\%             \\
                                    & \multicolumn{1}{c|}{\textbf{fiction}}    & 72.93\%       & 78.89\%        & 74.60\%             \\
                                    & \multicolumn{1}{c|}{\textbf{government}} & 76.76\%       & 83.48\%        & 77.00\%             \\
                                    & \multicolumn{1}{c|}{\textbf{letters}}    & 73.66\%       & 82.71\%        & 76.72\%             \\
                                    & \multicolumn{1}{c|}{\textbf{nineeleven}} & 76.01\%       & 82.51\%        & 76.82\%             \\
                                    & \multicolumn{1}{c|}{\textbf{oup}}        & 70.60\%       & 76.05\%        & 70.56\%             \\
                                    & \multicolumn{1}{c|}{\textbf{slate}}      & 68.94\%       & 75.21\%        & 68.49\%             \\
                                    & \multicolumn{1}{c|}{\textbf{telephone}}  & 70.67\%       & 77.69\%        & 71.97\%             \\
                                    & \multicolumn{1}{c|}{\textbf{travel}}     & 75.07\%       & 82.86\%        & 73.81\%             \\
                                    & \multicolumn{1}{c|}{\textbf{verbatim}}   & 68.11\%       & 72.20\%        & 68.29\%             \\ \bottomrule
\end{tabularx}%
}
\caption{Accuracy and f1-score values obtained with SILT grouped by genre of the text, only reported for XNLI. The genre represents several types of intentionality and style. Detailed descriptions can be found in the original MNLI article~\cite{Williams_Nangia_Bowman_2018}.}
\label{tab:3_results_by_genre_xnli}
\end{table}

Table \ref{tab:5_results_by_length} compares metrics by character length of sentences, to evaluate sensitivity to long sequences. SNLI always has shorter sequences, and on average are well classified across all lengths. There is also some mismatch between the XLM-R long sentence classification accuracy, dropping from 82.35\% accuracy to 56.62\% F1-score on SICK. On XNLI sentence length is more variable and there is always some decrease in accuracy on mid-range sequences. Using XLM-R as a base results in very high accuracy (94.41\%) for very long sequences, probably due to longer sentences containing more information about its contents. 

\begin{table}[htpb]
\centering
\resizebox{\textwidth}{!}{%
\begin{tabularx}{1.3\textwidth}{@{}YYYYYYY@{}}

\toprule
\multirow{2}{*}{\textbf{Metric}} & \multirow{2}{*}{\textbf{Dataset}} & \multirow{2}{*}{\textbf{\begin{tabular}[c]{@{}c@{}}Base\\ transformer\end{tabular}}} & \multicolumn{4}{c}{\textbf{Length range}} \\ \cmidrule(l){4-7} 
 &
   &
   &
  \textbf{(0, 125{]}} &
  \textbf{(125, 250{]}} &
  \textbf{(250, 375{]}} &
  \textbf{(375, 500{]}} \\ \midrule\midrule
\multirow{6}{*}{\textbf{Accuracy}} &
  \multirow{3}{*}{\textbf{SICK}} &
  \multicolumn{1}{c|}{\textbf{BERT}} &
  79.20\% &
  78.42\% &
  88.24\% &
  - \\
 &
   &
  \multicolumn{1}{c|}{\textbf{XLM-R}} &
  82.56\% &
  82.08\% &
  82.35\% &
  - \\
 &
   &
  \multicolumn{1}{c|}{\textbf{distilBERT}} &
  82.94\% &
  79.96\% &
  82.35\% &
  - \\ \cmidrule(l){2-7} 
 &
  \multirow{3}{*}{\textbf{XNLI}} &
  \multicolumn{1}{c|}{\textbf{BERT}} &
  73.03\% &
  72.13\% &
  75.44\% &
  75.00\% \\
 &
   &
  \multicolumn{1}{c|}{\textbf{XLM-R}} &
  79.40\% &
  78.78\% &
  81.77\% &
  96.43\% \\
 &
   &
  \multicolumn{1}{c|}{\textbf{distilBERT}} &
  74.95\% &
  72.72\% &
  74.31\% &
  78.57\% \\ \midrule
\multirow{6}{*}{\textbf{F1-Score}} &
  \multirow{3}{*}{\textbf{SICK}} &
  \multicolumn{1}{c|}{\textbf{BERT}} &
  76.41\% &
  71.86\% &
  88.19\% &
  - \\
 &
   &
  \multicolumn{1}{c|}{\textbf{XLM-R}} &
  81.87\% &
  79.62\% &
  56.62\% &
  - \\
 &
   &
  \multicolumn{1}{c|}{\textbf{distilBERT}} &
  82.53\% &
  77.05\% &
  82.11\% &
  - \\ \cmidrule(l){2-7} 
 &
  \multirow{3}{*}{\textbf{XNLI}} &
  \multicolumn{1}{c|}{\textbf{BERT}} &
  72.78\% &
  72.07\% &
  75.48\% &
  67.29\% \\
 &
   &
  \multicolumn{1}{c|}{\textbf{XLM-R}} &
  79.11\% &
  78.68\% &
  81.80\% &
  94.41\% \\
 &
   &
  \multicolumn{1}{c|}{\textbf{distilBERT}} &
  74.73\% &
  72.70\% &
  74.38\% &
  72.84\% \\ \bottomrule
\end{tabularx}%
}
\caption{Accuracy and f1-score grouped by semantic similarity, only reported for SICK. Contradiction and entailment only occur from 2 similarity points onwards, as such the first range goes from 0 to 3. Similarity ranges from 0 (unrelated) to 5 (exactly the same topic).}
\label{tab:5_results_by_length}
\end{table}

Finally we provide a qualitative comparison of our results in \ref{tab:8_qualitative_results}, showing the results of SILT with XML-R as the base transformer.

\begin{table}[htpb]
\centering
\resizebox{\textwidth}{!}{%
\begin{tabular}{@{}ccp{8.5cm}p{8.5cm}cc@{}}
\toprule
\multicolumn{1}{c}{\textbf{Data}} &
  \multicolumn{1}{c}{\textbf{Lan.}} &
  \multicolumn{1}{c}{\textbf{Sentence A}} &
  \multicolumn{1}{c}{\textbf{Sentence B}} &
  \multicolumn{1}{c}{\textbf{Obs.}} &
  \multicolumn{1}{c}{\textbf{Pred.}} \\ \midrule
\multirow{12}{*}{\textbf{XNLI}} &
  \multirow{3}{*}{\textbf{En-En}} &
  - There's some cash flow projections on my desk and, um, uh, it's for such and such Cutty, that's the client's name. &
  - We don't have any clients called Cutty. &
  Con. &
  Con. \\
 &
   &
  - But they were divided about like who were the field hands and who were the house kids, it was kind of-- &
  \begin{tabular}[c]{@{}l@{}}- They couldn't agree about who was a field\\ hand and who belonged in the house.\end{tabular} &
  Ent. &
  Ent. \\
 &
   &
  - And, of course, Androv Gromikov didn't answer anything, but we had all the information from the films the U2 had taken. &
  \begin{tabular}[c]{@{}l@{}}- The U2 took a ton of film from under the\\  water.\end{tabular} &
  Neu. &
  Con. \\ \cmidrule(l){2-6} 
 &
  \multirow{3}{*}{\textbf{En-Es}} &
  - There's some cash flow projections on my desk and, um, uh, it's for such and such Cutty, that's the client's name. &
  - No tenemos clientes llamados Cutty. &
  Con. &
  Con. \\
 &
   &
  - But they were divided about like who were the field hands and who were the house kids, it was kind of-- &
  - No podían ponerse de acuerdo sobre quién era una mano de campo y quién pertenecía a la casa. &
  Ent. &
  Ent. \\
 &
   &
  - And, of course, Androv Gromikov didn't answer anything, but we had all the information from the films the U2 had taken. &
  - El U2 tomó una tonelada de película de debajo del agua. &
  Neu. &
  Neu. \\ \cmidrule(l){2-6} 
 &
  \multirow{3}{*}{\textbf{Es-En}} &
  - Hay algunas proyecciones del flujo de caja y, uh, eh, es para tal y cual Cutty, el nombre del cliente. &
  - We don't have any clients called Cutty. &
  Con. &
  Con. \\
 &
   &
  - Pero ellos estaban divididos acerca de quiénes eran las manos de campo y quiénes eran los niños de la casa, era algo así como... &
  - They couldn't agree about who was a field hand and who belonged in the house. &
  Ent. &
  Ent. \\
 &
   &
  - Y, por supuesto, Androv Gromikov no respondió a nada, pero disponíamos de toda la información de las películas hechas por el U2. &
  - The U2 took a ton of film from under the water. &
  Neu. &
  Con. \\ \cmidrule(l){2-6} 
 &
  \multirow{3}{*}{\textbf{Es-Es}} &
  - Hay algunas proyecciones del flujo de caja y, uh, eh, es para tal y cual Cutty, el nombre del cliente. &
  - No tenemos clientes llamados Cutty. &
  Con. &
  Con. \\
 &
   &
  - Pero ellos estaban divididos acerca de quiénes eran las manos de campo y quiénes eran los niños de la casa, era algo así como... &
  - No podían ponerse de acuerdo sobre quién era una mano de campo y quién pertenecía a la casa. &
  Ent. &
  Ent. \\
 &
   &
  - Y, por supuesto, Androv Gromikov no respondió a nada, pero disponíamos de toda la información de las películas hechas por el U2. &
  - El U2 tomó una tonelada de película de debajo del agua. &
  Neu. &
  Neu. \\ \midrule
\multirow{12}{*}{\textbf{SICK}} &
  \multirow{3}{*}{\textbf{En-En}} &
  - Nobody is practicing water safety and wearing preservers &
  - This group of people is practicing water safety and wearing preservers &
  Con. &
  Con. \\
 &
   &
  - Two groups of people are playing football &
  - Two teams are competing in a football match &
  Ent. &
  Ent. \\
 &
   &
  - Two women are sparring in a kickboxing match &
  - Two people are fighting and spectators are watching &
  Neu. &
  Neu. \\ \cmidrule(l){2-6} 
 &
  \multirow{3}{*}{\textbf{En-Es}} &
  - Nobody is practicing water safety and wearing preservers &
  - Este grupo de personas practica la seguridad en el agua y usa preservativos &
  Con. &
  Con. \\
 &
   &
  - Two groups of people are playing football &
  - Dos equipos compiten en un partido de fútbol. &
  Ent. &
  Ent. \\
 &
   &
  - Two women are sparring in a kickboxing match &
  - Dos personas pelean y los espectadores miran &
  Neu. &
  Neu. \\ \cmidrule(l){2-6} 
 &
  \multirow{3}{*}{\textbf{Es-En}} &
  - Nadie practica la seguridad en el agua y usa preservativos &
  - This group of people is practicing water safety and wearing preservers &
  Con. &
  Con. \\
 &
   &
  - Dos grupos de personas están jugando al fútbol. &
  - Two teams are competing in a football match &
  Ent. &
  Ent. \\
 &
   &
  - Dos mujeres pelean en un combate de kickboxing &
  - Two people are fighting and spectators are watching &
  Neu. &
  Neu. \\ \cmidrule(l){2-6} 
 &
  \multirow{3}{*}{\textbf{Es-Es}} &
  - Nadie practica la seguridad en el agua y usa preservativos &
  - Este grupo de personas practica la seguridad en el agua y usa preservativos &
  Con. &
  Con. \\
 &
   &
  - Dos grupos de personas están jugando al fútbol. &
  - Dos equipos compiten en un partido de fútbol. &
  Ent. &
  Ent. \\
 &
   &
  - Dos mujeres pelean en un combate de kickboxing &
  - Dos personas pelean y los espectadores miran &
  Neu. &
  Neu. \\ \bottomrule
\end{tabular}%
}
\caption{Qualitative comparison of sentences, with their respective observation and predicted label. Predictions are made using the SILT with the XLM-R base.}
\label{tab:8_qualitative_results}
\end{table}

\section{Conclusions}
\label{sec:conclusiones}
Quality multilingual understanding of natural language is a resource-intensive task, where training and fine-tuning can be unwieldy. Our focus on an efficient training allows to reach state-of-the-art results with modest equipment, with the additional improvement of inter-lingual understanding. The SILT topology presented in this research consists of an architecture appended to frozen transformer models to understand entailment relationships between matched and mixed language sentence pairs. With a huge reduction of 92\% of trainable parameters, we achieve 82.5\% accuracy for SICK and 79.3\% accuracy for XNLI, both common benchmarks for NLI.

First, the inter-lingual capabilities of the model should be highlited. To the best knowledge of the authors, mixed language sentence pairs never share input space and the SILT topology succesfully bridges this gap. Despite drops in performance in comparison to matched language sentence pairs, inter-lingual results remain robust. In short, SILT is able to mix languages in its input stage.

Although comparisons against state-of-the-art models reveal that there is a noticeable drop in accuracy using SILT, showing an evident trade-off between training efficiency and accuracy, this drop is minimal when testing in the SICK dataset, maintaining high accuracy on Spanish sentence pairs and, more importantly, inter-lingual scenarios. As for XNLI, we observed that, despite the obvious drop in English accuracy, the cross-lingual sentence pairs are comparable to state of the art methods. To emphasize the contributions of SILT, all of these results are achieved with a fraction of trainable parameters of the original architecture and drastically reducing the computational resources required, from a distributed system of hundreds of GPUs to just one single GPU.

Throughout this research we have found that frozen transformers and a linear head are not as good as expected, which is evidenced by the shortcomings of the Sentence-BERT topology. If transformers are themselves few-show unsupervised learners, the representations they build should also be powerful even without usual fine-tuning. Instead, we find that these representations are insufficient to reliably detect entailment relationships between sentence pairs, either mono or inter-lingual. Besides, we suspect that fine tuning compromises multilingual capabilities for inter-lingual sentence pairs, and that this should be looked into. Comparing English to Spanish results, it is clear that the native language biases the accuracy of these models towards English, and that other languages are aligned and subject to it. This also occurs for SILT, where the frozen transformers also suffer from this effect but to a lesser degree, given that they have not been fine-tuned for the task at hand.

\section{Acknowledgements}
\label{sec:acknowledgements}
This research is funded by the project CIVIC: Intelligent characterisation of the veracity of the information related to COVID-19, granted by BBVA FOUNDATION GRANTS FOR SCIENTIFIC RESEARCH TEAMS SARS-CoV-2 and COVID-19.

This research has been supported by Ministry of Science and Education under TIN2017-85727-C4-3-P (DeepBio) project, by CHIST-ERA 2017 BDSI PACMEL Project (PCI2019-103623), and by Comunidad Aut\'{o}-noma de Madrid under S2018/ TCS-4566 (CYNAMON),  S2017/BMD-3688 grant. We gratefully acknowledge the support of NVIDIA Corporation with the donation of the Titan V GPU used for this research. 

This work was supported by the Comunidad de Madrid under Convenio Plurianual with the Universidad Politécnica de Madrid in the actuation line of Programa de Excelencia para el Profesorado Universitario”.

\bibliographystyle{unsrt}
\bibliography{bibliography.bib}

\end{document}